# DISTRIBUTED UNMIXING OF HYPERSPECTRAL DATA WITH SPARSITY CONSTRAINT

S. Khoshsokhan [a], R. Rajabi [a], H. Zayyani [a]

[a] Qom University of Technology, Electrical and Computer Engineering Department, Qom, Iran - (khoshsokhan.s, rajabi, zayyani)@qut.ac.ir

**ABSTRACT:**

Spectral unmixing (SU) is a data processing problem in hyperspectral remote sensing. The significant challenge in the SU problem is how to identify endmembers and their weights, accurately. For estimation of signature and fractional abundance matrices in a blind problem, nonnegative matrix factorization (NMF) and its developments are used widely in the SU problem. One of the constraints which was added to NMF is sparsity constraint that was regularized by $L_{1/2}$ norm. In this paper, a new algorithm based on distributed optimization has been used for spectral unmixing. In the proposed algorithm, a network including single-node clusters has been employed. Each pixel in hyperspectral images considered as a node in this network. The distributed unmixing with sparsity constraint has been optimized with diffusion LMS strategy, and then the update equations for fractional abundance and signature matrices are obtained. Simulation results based on defined performance metrics, illustrate advantage of the proposed algorithm in spectral unmixing of hyperspectral data compared with other methods. The results show that the AAD and SAD of the proposed approach are improved respectively about 6 and 27 percent toward distributed unmixing in SNR=25dB.

## 1. INTRODUCTION

Hyperspectral remote sensors capture the electromagnetic energy emitted from materials and collect hyperspectral images as a data cube with two-dimensional spatial information over many contiguous bands of high spectral resolution. One of the challenges in hyperspectral imaging is mixed pixels, which are pixels containing more than one kind of materials. So, spectrum of a pixel is a mixture of spectrum of some materials in the scene, named endmembers. Each endmember in a pixel is weighted by its fractional abundance (Miao and Qi, 2007). Decomposition of the mixed pixels is known as spectral unmixing (SU) problem (Mei et al., 2011). Most of the spectral unmixing methods are based on linear mixing model (LMM), in which it is assumed that the recorded spectrum of a particular pixel is linearly mixed by endmembers, which exist in the pixel. If the number of endmembers that are present in the scene and its signatures, are unknown, the SU problem becomes a blind source separation (BSS) problem (Qian et al., 2011).

Several SU methods were proposed in different models. Pixel purity index (PPI) (Chang and Plaza, 2006), N-FINDR (Winter, 1999), simplex volume maximization (Chan et al., 2011), convex cone analysis (CCA) (Ifarraguerri and Chang, 1999), successive projections algorithm (SPA) (Araújo et al., 2001), principal component analysis (PCA) (Smith et al., 1985), vertex component analysis (VCA) (Nascimento and Dias, 2005), (Lopez et al., 2012), are some of convex geometric methods. They are based on the pure pixel assumption, that the simplex volume is considered as a criterion for detection of endmembers. Some of the methods such as Independent Component Analysis (ICA) (Bayliss et al., 1998) use statistical models to solve the SU problem.

Another method is nonnegative matrix factorization (NMF) (Paatero and Tapper, 1994), (Lee and Seung, 1999), which decomposes the data into two nonnegative matrices. Recently, this basic method was developed with adding constraints, such as the minimum volume constrained NMF (MVC-NMF) method (Miao and Qi, 2007), graph regularized NMF (GNMF) (Rajabi and Ghassemian, 2013), NMF with local smoothness constraint (NMF-LSC) (Yang et al., 2015), multilayer NMF (MLNMF) (Rajabi and Ghassemian, 2015), structured discriminative NMF (SDNMF) (Li et al., 2016), and region-based structure preserving NMF (R-NMF) (Tong et al., 2017). One of the constraints that has been used to improve performance of NMF methods is sparsity constraint that can be applied to NMF cost function using $L_q$ regularizers. The sparsity constraint means that most of the pixels composed of only a few of the endmembers in the scene (Iordache et al., 2010), and the fractional abundances of other endmembers are equal to zero. So, the abundance matrix has many zero elements, and it has a large degree of sparsity. Regularization methods have been used to provide updating equations for signatures and abundances. Using $L_{1/2}$ regularization into NMF, which leads to an algorithm named $L_{1/2}$-NMF, has been proposed in (Qian et al., 2011), that enforces the sparsity of fractional abundances.

As another approach, the distributed strategy has been used for utilization of neighborhood information. There are some distributed strategies such as consensus strategies (Tsitsiklis and Athans, 1984), incremental strategies (Bertsekas, 1997) and diffusion strategies (Sayed et al., 2013). In this article, a diffusion strategy is used because it has high stability over adaptive networks (Sayed et al., 2013). Diffusion LMS strategy has been proposed in (Cattivelli and Sayed, 2010).

To solve a distributed problem, a network is considered. There are three types of networks: 1) a single-task network, that nodes estimate a common unknown and optimum vector, 2) a multitask network, which each node estimate its own optimum vector and 3) a clustered multitask network includes clusters that each of them has to estimate a common optimum vector (Chen et al., 2014). Unmixing problem is a multitask problem where each pixel considered to be as a node. Here, we first consider the general case, where there is a clustered multitask network and each cluster has an optimum vector (fractional abundance vector) that should be estimated. Then we will reduce that to a multitask case.

In this paper, the distributed unmixing of hyperspectral images is considered in which neighborhood information and sparsity constraint are used.

This paper is organized as follows. Section 2 presents the proposed method and optimization procedure. Section 3 includes introduction of datasets. Section 4 provides simulation results and the last section gives conclusions.

## 2. DISTRIBUTED UNMIXING OF HYPERSPECTRAL DATA WITH SPARSITY CONSTRAINT

In this section, a new method that utilizes sparsity constraint and neighborhood information is proposed. First, we will express linear mixing model in subsection 2.1, then we will formulate the distributed cost functions with sparsity constraint in 2.2, and finally, we will use them to solve SU problem in 2.3.

### 2.1 Linear Mixing Model (LMM)

To solve the SU problem, we focus on a simple but efficient model, named LMM. In this model, there exists a linear relation between the endmembers that weighted by their fractional abundances, in the scene. Mathematically, this model is described as:

$$\mathbf{y} = \mathbf{A}\mathbf{s} + \boldsymbol{\varepsilon} \tag{1}$$

where $\mathbf{y}$ is an observed data vector, $\mathbf{A}$ is the signature matrix, $\mathbf{s}$ is the fractional abundance vector and $\boldsymbol{\varepsilon}$ is assumed as a noise vector.

In the SU problem, fractional abundance vectors have two constraints in each pixel, abundance sum to one constraint (ASC) and abundance nonnegativity constraint (ANC) (Ma et al., 2014), which are as follows, for $p$ endmembers in a scene.

$$\sum_{n=1}^{p} \mathbf{s}_k(n) = 1 \tag{2}$$

$$\mathbf{s}_k(n) > 0, n = 1,..,p \tag{3}$$

Where $\mathbf{s}_k(n)$ is the fractional abundance of the $n$-th endmember in the $k$-th pixel of the image. Some methods have been proposed without applying ASC constraint. For instance, this constraint has not been used in (Zhang et al., 2012) for complicated ground scene with nonlinear interferences. However, according to (Heinz, 2001), that studied utilization or removing of this constraint, in this article ASC constraint is adopted to gain better results.

### 2.2 Distributed Cost Functions and Optimization

As explained earlier, three types of networks containing single task, multitask and clustered multitask networks are supposed. First, $N$ nodes are considered in a clustered multitask network as pixels in a hyperspectral image, and a $p \times 1$ optimum vector $\mathbf{s}_{C(k)}$ is estimated, and $\mathbf{a}_k$ is the signature vector at node $k$. A cost function $J_k(\mathbf{s}_{C(k)})$, when node $k$ is in cluster $C(k)$, defined as follows from the mean square error criterion (Chen et al., 2014):

$$J_k(\mathbf{s}_{C(k)}) = E\left\{|\mathbf{y}_k(n) - \mathbf{a}_k^T(n)\mathbf{s}_{C(k)}|^2\right\} \tag{4}$$

Then, the following equation is written, using the iterative steepest-descent solution (Sayed, 2003):

$$\mathbf{s}_{C(k)}(n) = \mathbf{s}_{C(k)}(n-1) - \mu[\nabla_s J(\mathbf{s}_{C(k)}(n-1))]^* \tag{5}$$

where $\mu > 0$ is a step-size parameter, then by computing complex gradient and substituting it into (5), the following iterative equation is obtained:

$$\mathbf{s}_{C(k)}(n) = \mathbf{s}_{C(k)}(n-1) + \mu \sum_{k=1}^{N} \left( R_{\mathbf{ya},k} - R_{\mathbf{a},k} \mathbf{s}_{C(k)}(n-1) \right) \tag{6}$$

This equation requires knowledge of the autocorrelations, and they are replaced by instantaneous approximations, in the LMS algorithm as follows:

$$R_{\mathbf{a},k} \approx \mathbf{a}_k^*(n)\mathbf{a}_k(n) \tag{7}$$

$$R_{\mathbf{ya},k} \approx \mathbf{y}_k(n)\mathbf{a}_k^*(n) \tag{8}$$

Then, the recursive equation is changed to:

$$\mathbf{s}_{C(k)}(n) = \mathbf{s}_{C(k)}(n-1) + \mu \sum_{k=1}^{N} \mathbf{a}_k^*(n)[\mathbf{y}_k(n) - \mathbf{a}_k(n)\mathbf{s}_{C(k)}(n-1)] \tag{9}$$

Equation (9) is not distributed, because it requires knowledge of $\{\mathbf{y}_k, \mathbf{a}_k\}$ from the entire network (Cattivelli and Sayed, 2010).

In a distributed network, relationships between nodes are used for improving the accuracy. In this article, we utilize the squared Euclidean distance (Chen et al., 2014):

$$\Delta(\mathbf{s}_{C(k)}, \mathbf{s}_{C(\ell)}) = \|\mathbf{s}_{C(k)} - \mathbf{s}_{C(\ell)}\|^2 \tag{10}$$

Then, the $L_{1/2}$ regularizer for sparsity constraint is used (Qian et al., 2011):

$$\|\mathbf{s}\|_{1/2} = \sum_{k,n=1}^{N,p} \mathbf{s}_k^{1/2}(n) \tag{11}$$

Combining (4), (10) and (11), the following cost function is obtained:

$$J(\mathbf{s}_{C_1},...,\mathbf{s}_{C_Q}) = \sum_{k=1}^{N} E\{|\mathbf{y}_k(n) - \mathbf{a}_k^T(n)\mathbf{s}_{C(k)}|^2\} \\ + \eta \sum_{k=1}^{N} \sum_{l \in N_k \setminus C(k)} \rho_{kl} \|\mathbf{s}_{C(k)} - \mathbf{s}_{C(l)}\|^2 + \lambda \sum_{k,n=1}^{N,p} \mathbf{s}_k^{1/2}(n) \tag{12}$$

where it is the cost function for abundances of $Q$ clusters, and the symbol \ is the set difference, $\eta > 0$ is a regularization parameter (Chen et al., 2014) that controls the effect of neighborhood term, $\lambda$ is a scalar that weights the sparsity function (Qian et al., 2011), $N_k$ shows nodes that are in the neighborhood of node $k$, and the nonnegative coefficients $\rho_{kl}$ are normalized spectral similarity which is obtained from correlation of data vectors (Chen et al., 2014). The coefficients are computed as introduced in (Qian et al., 2011), (Chen et al., 2014):

$$\lambda = \frac{1}{\sqrt{L}} \sum_l \frac{\sqrt{N} - \|\mathbf{y}_l\|_1 / \|\mathbf{y}_l\|_2}{\sqrt{N}-1} \quad (13)$$

$$\rho_{kj} = \frac{\theta(\mathbf{y}_k, \mathbf{y}_j)}{\sum_{l \in N_k^-} \theta(\mathbf{y}_k, \mathbf{y}_l)} \quad (14)$$

where $N_k^-$ includes neighbors of node k except itself, and $\theta$ is computed as (Chen et al., 2014):

$$\theta(\mathbf{y}_k, \mathbf{y}_j) = \frac{\mathbf{y}_k^T \mathbf{y}_j}{\|\mathbf{y}_k\| \|\mathbf{y}_j\|} \quad (15)$$

If $\mathbf{s}_k^o$ is considered as the minimizer of the cost function, this equation is denoted as:

$$J_k(\mathbf{s}_k) = J_k(\mathbf{s}_k^o) + \|\mathbf{s}_k - \mathbf{s}_k^o\|_{R_k}^2 \quad (16)$$

Rayleigh-Ritz characterization eigenvalues (Sayed, 2013), is a strategy that let us to simplify the second term of equation (16) to:

$$\|\mathbf{s}_k - \mathbf{s}_l^o\|_{R_l}^2 \approx b_{lk} \|\mathbf{s}_k - \mathbf{s}_l^o\|^2 \quad (17)$$

where $b_{lk}$ are some nonnegative scalar (Chen et al., 2014). Then the cost function changes as follows for one cluster:

$$J_{C(k)}(\mathbf{s}_k) = \sum_{l \in N_k \cap C(k)} E\{|\mathbf{y}_l(n) - \mathbf{a}_l^T(n)\mathbf{s}_k|^2\}$$
$$+ \eta \sum_{l \in N_k \setminus C(k)} \rho_{kl} \|\mathbf{s}_k - \mathbf{s}_l\|^2 + \lambda \mathbf{s}_k^{1/2}(n) \quad (18)$$
$$+ \sum_{l \in N_k^- \cap C(k)} b_{lk} \|\mathbf{s}_k - \mathbf{s}_l^o\|^2$$

Now, minimizing this cost function, using steepest-descent in equation (5), results to:

$$\mathbf{s}_k(n+1) = \mathbf{s}_k(n) - \mu \sum_{l \in N_k \cap C(k)} (R_{a,l}\mathbf{s}_k(n) - R_{ya,k})$$
$$- \mu\eta \sum_{l \in N_k \setminus C(k)} \rho_{kl}(\mathbf{s}_k(n) - \mathbf{s}_l(n)) - \mu\lambda \mathbf{s}_k^{-1/2}(n) \quad (19)$$
$$- \mu \sum_{l \in N_k^- \cap C(k)} b_{lk}(\mathbf{s}_k(n) - \mathbf{s}_l^o)$$

As explained earlier, the SU is a multitask problem that each cluster only has one node. Thus, above equation with adoption of LMS strategy, is simplified to:

$$\mathbf{s}_k(n+1) = \mathbf{s}_k(n) + \mu[\mathbf{y}_k(n) - \mathbf{a}_k^T(n)\mathbf{s}_k(n)]\mathbf{a}_k(n)$$
$$+ \eta\mu \sum_{l \in N_k^-} \rho_{kl}(\mathbf{s}_l(n) - \mathbf{s}_k(n)) - \mu\lambda \mathbf{s}_k^{-1/2}(n) \quad (20)$$

Therefore, the optimum vectors are computed with the recursive equation, using some initial values.

### 2.3 Spectral Unmixing Updating Equations

According to the NMF algorithm, the conventional least squares error should be minimized with respect to the signatures and abundances matrices, subject to the non-negativity constraint (Lee and Seung, 2001). So, we have:

$$\min_{\mathbf{S},\mathbf{A} > 0} \|\mathbf{Y} - \mathbf{AS}\|_F^2 \quad (21)$$

where $\mathbf{A}$ and $\mathbf{S}$ are signatures and fractional abundances matrices, respectively, and $Y$ denotes Hyperspectral data matrix. Then, similar to the distributed unmixing with sparsity constraint, these terms are added to the NMF cost function as follows:

$$\min_{\mathbf{S},\mathbf{A} > 0} \|\mathbf{Y} - \mathbf{AS}\|_F^2 + \eta \sum_{k=1}^{N} \sum_{j \in N_k} \rho_{kj} \|\mathbf{s}_k - \mathbf{s}_j\|_1$$
$$+ \lambda \sum_{k=1}^{N} \mathbf{s}_k^{1/2}(n) \quad (22)$$

This cost function has been minimized with respect to $\mathbf{A}$, using multiplicative update rules (Lee and Seung, 2001), and then recursive equation for signatures matrix is obtained as:

$$\mathbf{A} = \mathbf{A} \cdot \frac{\mathbf{YS}^T}{\mathbf{ASS}^T} \quad (23)$$

Then, according to equation (22), and using $d\|x\|_1/dx = \text{sign}(x)$, recursive equation for abundance vectors is obtained as follows:

$$\mathbf{s}_k(n+1) = \mathbf{s}_k(n) + \mu\mathbf{A}^T(\mathbf{y}_k - \mathbf{As}_k(n))$$
$$- \mu\eta \sum_{j \in N_k} \rho_{kj} \text{sign}(\mathbf{s}_k(n) - \mathbf{s}_j(n)) \quad (24)$$
$$- \mu\lambda \mathbf{s}_k^{-1/2}(n)$$

In this paper, the ASC and ANC constraints are adopted for abundance vectors, with using the operator that explained in (Chen and Ye, 2011). This operator projects vectors onto a simplex which size of its sides are equal to one.

## 3. DATASETS

The proposed algorithm is tested on synthetic and real data. This section introduces a real dataset that recorded with hyperspectral sensors and a synthetic data set that are generated using spectral libraries.

### 3.1 Synthetic Images

To generate synthetic data, some spectral signatures are chosen from a digital spectral library (USGS) (Clark et al., 2007), that include 224 spectral bands, with wavelengths from 0.38μm to 2.5μm. Size of intended images is 64×64, and one endmember has been contributed in spectral signature of each pixel, randomly. Pixels of this image are pure, so to have an image containing mixed pixels, a low pass filter is considered. It averages from abundances of endmembers in its window, so that the LMM would be confirmed. Size of the window controls degree of mixing (Miao and Qi, 2007). With smaller dimension of the window and more endmembers in the image, degree of sparsity is increased.

## 3.2 Real Data

The real dataset that the proposed method was applied on it, is hyperspectral data captured by the Airborne Visible/Infrared Imaging Spectrometer (AVIRIS) over Cuprite, Nevada. This dataset has been used since the 1980s. AVIRIS spectrometer has 224 channels and covers wavelengths from 0.4μm to 2.5μm. Its spectral and spatial resolutions is about 10nm and 20m, respectively (Green et al., 1998). 188 bands of these 224 bands are used in the experiments. The other bands (covering bands 1, 2, 104-113, 148-167, and 221-224) have been removed which are related to water-vapor absorption or low SNR bands. Figure 1 illustrates a sample band (band #3) of this dataset.

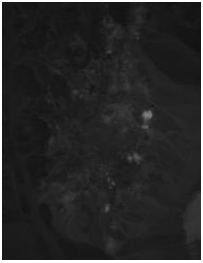

Figure 1. Band 3 of real data scene.

## 4. EXPERIMENTS AND RESULTS

In this section, for quantity comparison between the proposed and the other methods, the performance metrics such as spectral angle distance (SAD) and abundance angle distance (AAD) (Miao and Qi, 2007) are used. They are defined as:

$$SAD = \cos^{-1}\left(\frac{\mathbf{a}^T \hat{\mathbf{a}}}{\|\mathbf{a}\| \|\hat{\mathbf{a}}\|}\right) \quad (24)$$

$$AAD = \cos^{-1}\left(\frac{\mathbf{s}^T \hat{\mathbf{s}}}{\|\mathbf{s}\| \|\hat{\mathbf{s}}\|}\right) \quad (25)$$

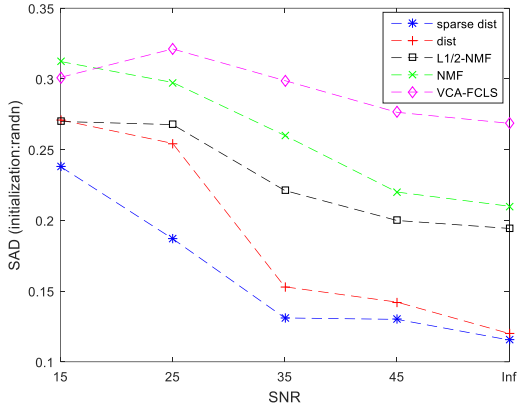

Figure 2. The SAD performance metric of 5 methods versus SNR, using random initialization and applied on synthetic data. SAD of the proposed algorithm is star-dashed line.

Table 1. Comparison of running time between four algorithms, using VCA initialization and SNR=25dB.

| method | Running time (second) |
| --- | --- |
| NMF | 33.5161 |
| $L_q$-NMF | 10.8671 |
| Distributed | 104.5395 |
| Sparse Distributed | 77.4153 |

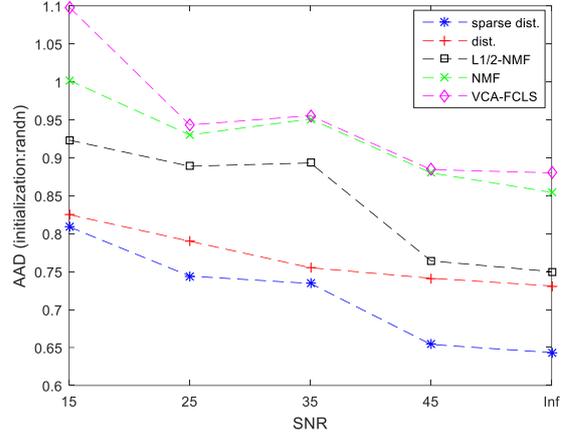

Figure 3. The AAD performance metric of 5 methods versus SNR, using random initialization and applied on synthetic data. AAD of the proposed algorithm is star-dashed line.

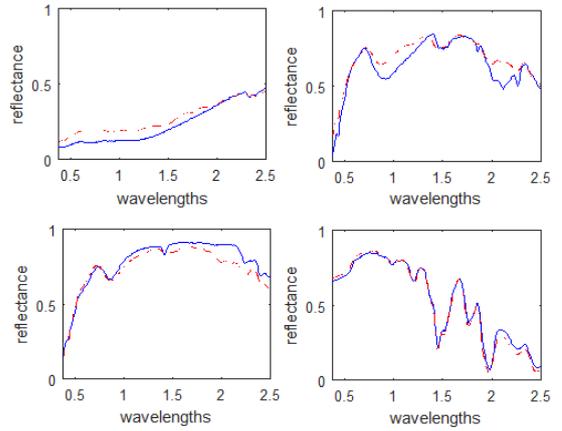

Figure 4. Original spectral signatures (blue solid lines) and estimated signatures using distributed unmixing with sparsity constraint (red dashed lines) versus wavelengths (μm), on synthetic data.

where $\hat{\mathbf{a}}$ is the estimation of spectral signature vectors and $\hat{\mathbf{s}}$ is the estimation of abundance fraction vectors.

The proposed algorithm and some other algorithms such as VCA-FCLS, NMF, $L_{1/2}$-NMF and distributed unmixing is applied on synthetic data, using a 3×3 low pass filter, with 4 endmembers, and without pure pixel. According to (Chen et al., 2014), this algorithm gain the best results with $(\mu,\eta) = (0.01,0.1)$. After generating data, noise is added to it with 5 different values of SNR, and then performance metrics are computed by averaging 20 Monte-Carlo runs. The simulation results are shown in figures 2, 3 and 4. Also Table 1 describe the average of running time of NMF, $L_q$-NMF, distributed unmixing and proposed method using MATLAB R2015b with Intel Core i5 CPU at 2.40 GHz and 4 GB memory. This table shows that one of the main advantages of sparse representation is its efficiency and improvement in running time. Afterwards, the proposed algorithm is applied on real data and simulation results are shown in figures 5 and 6. According to figures 2 and 3, the AAD and SAD measures decrease as the SNR values increase. The methods that eventuate lower level of SAD or AAD, have better performance. So the star-dashed lines illustrate that the proposed algorithm produces the best results in comparison with other algorithms.

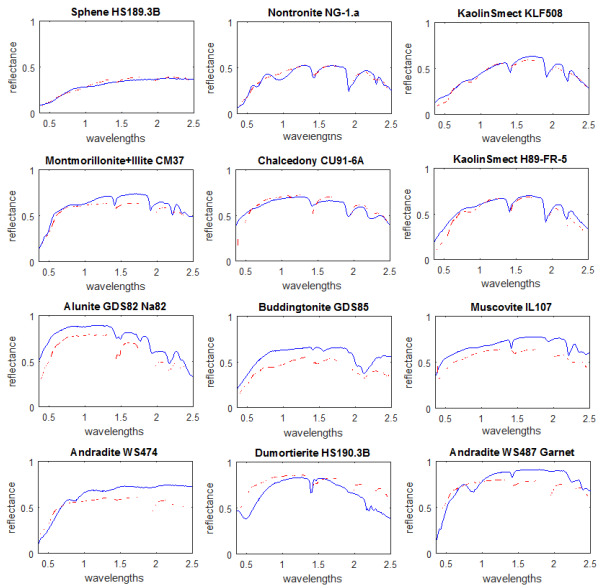

Figure 5. Original spectral signatures (blue solid lines) and estimated signatures using distributed unmixing with sparsity constraint (red dashed lines) versus wavelengths (μm), on AVIRIS dataset.

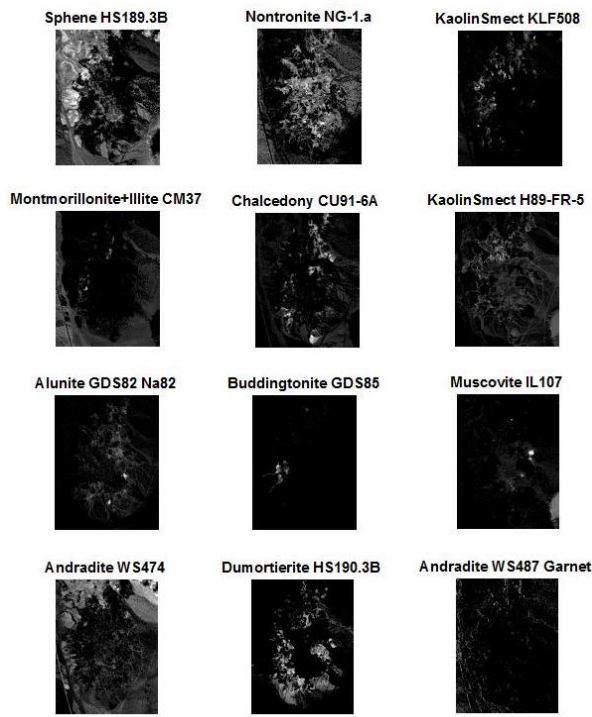

Figure 6. Fractional abundances of endmembers that are present in the scene.

## 5. CONCLUSION

Hyperspectral remote sensing is a distinguished research topic in data processing. The purpose of spectral unmixing is decomposition of pixels in the scene into their constituent materials. The proposed distributed unmixing method with sparsity constraint was developed that estimates signatures and their abundances, and improves fractional abundances estimation toward NMF method. Simulation results on real and synthetic dataset illustrated better performance of the proposed algorithm compared with NMF, $L_{1/2}$-NMF, VCA and distributed unmixing. Furthermore, this method converges faster in comparison with the distributed method.

## REFERENCES

Araújo, M.C.U., Saldanha, T.C.B., Galvao, R.K.H., Yoneyama, T., Chame, H.C., Visani, V., 2001. The successive projections algorithm for variable selection in spectroscopic multicomponent analysis. Chemometrics and Intelligent Laboratory Systems 57, 65-73.

Bayliss, J.D., Gualtieri, J.A., Cromp, R.F., 1998. Analyzing hyperspectral data with independent component analysis, 26th AIPR Workshop: Exploiting New Image Sources and Sensors. International Society for Optics and Photonics, pp. 133-143.

Bertsekas, D.P., 1997. A new class of incremental gradient methods for least squares problems. SIAM Journal on Optimization 7, 913-926.

Cattivelli, F.S., Sayed, A.H., 2010. Diffusion LMS strategies for distributed estimation. IEEE Transactions on Signal Processing 58, 1035-1048.

Chan, T.-H., Ma, W.-K., Ambikapathi, A., Chi, C.-Y., 2011. A simplex volume maximization framework for hyperspectral endmember extraction. IEEE Transactions on Geoscience and Remote Sensing 49, 4177-4193.

Chang, C.-I., Plaza, A., 2006. A fast iterative algorithm for implementation of pixel purity index. IEEE Geoscience and Remote Sensing Letters 3, 63-67.

Chen, J., Richard, C., Sayed, A.H., 2014. Multitask diffusion adaptation over networks. IEEE Transactions on Signal Processing 62, 4129-4144.

Chen, Y., Ye, X., 2011. Projection onto a simplex. arXiv preprint arXiv:1101.6081.

Clark, R.N., Swayze, G.A., Wise, R., Livo, K.E., Hoefen, T., Kokaly, R.F., Sutley, S.J., 2007. USGS digital spectral library splib06a. US Geological Survey, Digital Data Series 231.

Green, R.O., Eastwood, M.L., Sarture, C.M., Chrien, T.G., Aronsson, M., Chippendale, B.J., Faust, J.A., Pavri, B.E., Chovit, C.J., Solis, M., 1998. Imaging spectroscopy and the airborne visible/infrared imaging spectrometer (AVIRIS). Remote Sensing of Environment 65, 227-248.

Heinz, D.C., 2001. Fully constrained least squares linear spectral mixture analysis method for material quantification in hyperspectral imagery. IEEE transactions on geoscience and remote sensing 39, 529-545.

Ifarraguerri, A., Chang, C.-I., 1999. Multispectral and hyperspectral image analysis with convex cones. IEEE Transactions on Geoscience and Remote Sensing 37, 756-770.

Iordache, M.-D., Plaza, A., Bioucas-Dias, J., 2010. Recent developments in sparse hyperspectral unmixing, Geoscience

and Remote Sensing Symposium (IGARSS), 2010 IEEE International. IEEE, pp. 1281-1284.

Lee, D.D., Seung, H.S., 1999. Learning the parts of objects by non-negative matrix factorization. Nature 401, 788-791.

Lee, D.D., Seung, H.S., 2001. Algorithms for non-negative matrix factorization, Advances in neural information processing systems, pp. 556-562.

Li, X., Zhou, J., Tong, L., Yu, X., Guo, J., Zhao, C., 2016. Structured Discriminative Nonnegative Matrix Factorization for hyperspectral unmixing, IEEE International Conference on Image Processing (ICIP) 2016. IEEE, pp. 1848-1852.

Lopez, S., Horstrand, P., Callico, G.M., Lopez, J.F., Sarmiento, R., 2012. A novel architecture for hyperspectral endmember extraction by means of the modified vertex component analysis (MVCA) algorithm. IEEE Journal of Selected Topics in Applied Earth Observations and Remote Sensing 5, 1837-1848.

Ma, W.-K., Bioucas-Dias, J.M., Chan, T.-H., Gillis, N., Gader, P., Plaza, A.J., Ambikapathi, A., Chi, C.-Y., 2014. A signal processing perspective on hyperspectral unmixing: Insights from remote sensing. IEEE Signal Processing Magazine 31, 67-81.

Mei, S., He, M., Zhang, Y., Wang, Z., Feng, D., 2011. Improving spatial–spectral endmember extraction in the presence of anomalous ground objects. IEEE Transactions on Geoscience and Remote Sensing 49, 4210-4222.

Miao, L., Qi, H., 2007. Endmember extraction from highly mixed data using minimum volume constrained nonnegative matrix factorization. IEEE Transactions on Geoscience and Remote Sensing 45, 765-777.

Nascimento, J.M., Dias, J.M., 2005. Vertex component analysis: A fast algorithm to unmix hyperspectral data. IEEE transactions on Geoscience and Remote Sensing 43, 898-910.

Paatero, P., Tapper, U., 1994. Positive matrix factorization: A non-negative factor model with optimal utilization of error estimates of data values. Environmetrics 5, 111-126.

Qian, Y., Jia, S., Zhou, J., Robles-Kelly, A., 2011. Hyperspectral unmixing via $L_{1/2}$ sparsity-constrained nonnegative matrix factorization. IEEE Transactions on Geoscience and Remote Sensing 49, 4282-4297.

Rajabi, R., Ghassemian, H., 2013. Hyperspectral data unmixing using GNMF method and sparseness constraint, IEEE International Geoscience and Remote Sensing Symposium (IGARSS). IEEE, pp. 1450-1453.

Rajabi, R., Ghassemian, H., 2015. Spectral unmixing of hyperspectral imagery using multilayer NMF. IEEE Geoscience and Remote Sensing Letters 12, 38-42.

Sayed, A.H., 2003. Fundamentals of adaptive filtering. John Wiley & Sons.

Sayed, A.H., 2013. Diffusion adaptation over networks. Academic Press Library in Signal Processing.

Sayed, A.H., Tu, S.-Y., Chen, J., Zhao, X., Towfic, Z.J., 2013. Diffusion strategies for adaptation and learning over networks: an examination of distributed strategies and network behavior. IEEE Signal Processing Magazine 30, 155-171.

Smith, M.O., Johnson, P.E., Adams, J.B., 1985. Quantitative determination of mineral types and abundances from reflectance spectra using principal components analysis. Journal of Geophysical Research: Solid Earth 90.

Tong, L., Zhou, J., Li, X., Qian, Y., Gao, Y., 2017. Region-Based Structure Preserving Nonnegative Matrix Factorization for Hyperspectral Unmixing. IEEE Journal of Selected Topics in Applied Earth Observations and Remote Sensing 10, 1575-1588.

Tsitsiklis, J., Athans, M., 1984. Convergence and asymptotic agreement in distributed decision problems. IEEE Transactions on Automatic Control 29, 42-50.

Winter, M.E., 1999. N-FINDR: An algorithm for fast autonomous spectral end-member determination in hyperspectral data, SPIE's International Symposium on Optical Science, Engineering, and Instrumentation. International Society for Optics and Photonics, pp. 266-275.

Yang, Z., Yang, L., Cai, Z., Xiang, Y., 2015. Spectral unmixing based on nonnegative matrix factorization with local smoothness constraint, Signal and Information Processing (ChinaSIP), 2015 IEEE China Summit and International Conference on. IEEE, pp. 635-638.

Zhang, Y., Fan, X., Zhang, Y., Wei, R., 2012. Linear spectral unmixing with generalized constraint for hyperspectral imagery, Geoscience and Remote Sensing Symposium (IGARSS), 2012 IEEE International. IEEE, pp. 4106-4109.